# RESEARCH ON DUAL CHANNEL NEWS HEADLINE CLASSIFICATION BASED ON ERNIE PRE-TRAINING MODEL

Junjie Li and Hui Cao

Key Laboratory of China's Ethnic Languages and Information Technology of Ministry of Education, Northwest Minzu University Lanzhou, China

## ABSTRACT

*The classification of news headlines is an important direction in the field of NLP, and its data has the characteristics of compactness, uniqueness and various forms. Aiming at the problem that the traditional neural network model cannot adequately capture the underlying feature information of the data and cannot jointly extract key global features and deep local features, a dual-channel network model DC-EBAD based on the ERNIE pre-training model is proposed. Use ERNIE to extract the lexical, semantic and contextual feature information at the bottom of the text, generate dynamic word vector representations fused with context, and then use the BiLSTM-AT network channel to secondary extract the global features of the data and use the attention mechanism to give key parts higher The weight of the DPCNN channel is used to overcome the long-distance text dependence problem and obtain deep local features. The local and global feature vectors are spliced, and finally passed to the fully connected layer, and the final classification result is output through Softmax. The experimental results show that the proposed model improves the accuracy, precision and F1-score of news headline classification compared with the traditional neural network model and the single-channel model under the same conditions. It can be seen that it can perform well in the multi-classification application of news headline text under large data volume.*

## 1. INTRODUCTION

The essence of news headline classification is text classification, which aims to classify the target text data accurately. In today's era of exponential growth of data information, online news has become the most important and efficient carrier of social public information and even international information dissemination. As the core of news headlines, it is the finishing touch. A vivid and unique headline can quickly attract readers' attention and its authenticity can optimize the network information environment to a certain extent, guide public opinion, and convey the correct value orientation [1].

However, in the era of big data, information is complicated and confusing, news is being generated all the time, and news headlines are constantly increasing. All kinds of news data are efficiently processed, and the use of deep learning to identify and classify news headlines has gradually become a hot topic. Although traditional neural network models such as LSTM, CNN, BiLSTM, and DPCNN can better classify news headlines, they are not enough, and there is still room for improvement. In recent years, pre-training models such as ERNIE have emerged rapidly, which can extract the underlying lexical, semantic, and contextual information of text data, and generate corresponding dynamic word vector representations that integrate contextual context for the input text data. Therefore, the new trend is to combine the pre-trained language model and the deep neural network model, so as to better perform word vector representation and corresponding feature extraction on text data. However, the single-channel network model generally follows the fixed process of the word vector representation of the input text and the feature extraction. Therefore, in order to better perform feature extraction in different ranges and splicing representation of different learned features, a dual-channel neural network model DC-EBAD based on ERNIE pre-trained language model is proposed. The ERNIE pre-training model is used to represent the input text data with dynamic word vector representation of the context, and the BiLSTM-AT channel is used to extract global features and use the attention mechanism to give higher weights to key parts, and the deep local features extracted by DPCNN are spliced with it to form the final feature vector, then the learning of features in different ranges is completed, so as to achieve the propose of extracting and learning features in an all-round way.

## 2. RELATED WORK

In machine learning, natural language processing tasks focus on feature selection and feature extraction of input data. After learning and training, the final classification result is obtained through the classifier. Li X et al.[2] used term frequency-inverse document frequency (TF-IDF) to calculate the feature weights of keywords, and carried out weighting processing on the naive Bayes algorithm to realize the classification of Internet hot news text data. . In order to strengthen the feature expression of the input text vector, the pre-training model has become the mainstream of vector feature expression. Yuejun X et al. [3] used BERT pre-training to fuse the trained sentence vector with the continuously updated proper noun vector and assign the TF-IDF value of the proper noun as the weight to the vector, which solved the existing patent classification. The method is limited by the problem of unregistered words in the patent text. Chengyu Q et al.[4] proposed a semi-supervised classification method based on graph convolutional neural network (BD-GCN) in response to the problems of lack of annotations in online bidding documents, sparse text semantics, diverse data sources, and complex information structure. Information extraction technology constructs bidding document data into a special knowledge graph model and integrates external text information, and uses graph convolutional neural network to realize semi-supervised classification of bidding documents. Zemin H et al. [5] proposed a text sentiment classification model combining BERT and BiSRU-AT, using BERT to obtain dynamic word vector representations, using BiSRU to extract semantic features and context information twice, and then integrating the Attention mechanism to assign weights to the output to solve the traditional semantics. The model cannot solve the problems of polysemous word representation and the existing sentiment analysis model cannot fully capture long-distance semantic information. Di W et al.[6] proposed an ELMo-CNN-BiGRU dual-channel text sentiment classification method, which uses the ELMo and Glove pre-training models to generate dynamic and static word vectors respectively and stack embedding to generate input vectors, constructing a fusion convolutional neural network(CNN) and bidirectional gated recurrent unit (BiGRU) two-channel neural network model to obtain the local and global features of the text. Keming C et al.[7] classify ANN and text, construct vector space to describe text and extract features of different types of text through text segmentation and word frequency statistics, and use ANN features for learning to complete text classification tasks. The globalization environment is increasingly demanding natural speech processing tasks in small languages. HossainMd. R et al.[8] proposed intelligent text classification models including GloVe embedding and ultra-deep convolutional neural network (VDCNN) classifiers and embedding parameter recognition ( EPI) algorithm to select the best embedding parameters for low-resource languages, so that resource-constrained languages can be better used for natural language processing tasks. In summary, the most critical step for natural language processing tasks is the extraction of features, which directly affects the quality of text classification results.

On this basis, for network news headlines that are short and succinct and have different styles of data, a dual-channel model (Dual-Channel ERNIE -BiLSTM with Attention and DPCNN, DC-EBAD), through ERNIE for the underlying text data to perform dynamic word vector representation that contains lexical, semantic and contextual information as the input of the subsequent network channel, BiLSTM overcomes the inability of a one-way LSTM network to reverse Extract the shortcomings of the global semantic information of the text, and use the attention mechanism to give higher weight to the key parts of the global features of the bi-directional text sequence vector extracted by BiLSTM to strengthen the classification feature expression; at the same time, the deep pyramid convolutional neural network model is built to overcome the traditional volume The product neural network CNN cannot obtain the long-distance dependence of the text through convolution, and performs deep-level local feature extraction while greatly saving calculation time. Through the splicing of the output feature vector of the dual-channel model as the input of the Softmax fully connected layer classification, the news headline classification result is finally obtained. The results show that it has a higher accuracy rate than the set comparison model.

## 3. DC-EBAD DUAL-CHANNEL MODEL

Based on the ERNIE pre-training model combined with the attention mechanism of the bidirectional long-term short-term memory network channel and the deep pyramid convolutional network dual-channel model DC-EBAD structure is shown in the following figure.

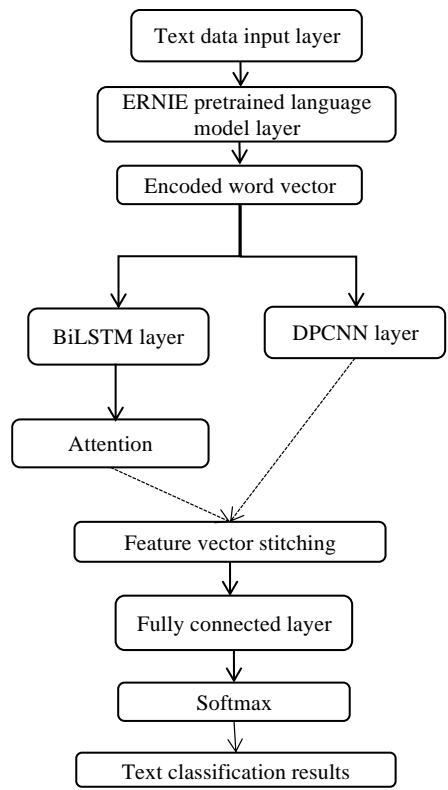

Figure 1. DC-EBAD network model structure

As shown in Figure.1 above, the input of the DC-EBAD neural network model is news headline text data, which is entered as headline short sentences. In the pre-training language model layer, it first passes through the internal word embedding layer to perform static word vector encoding in units of words. Then pass it to the ERNIE layer, and finally generate the corresponding dynamic word vector representation combined with the context. Then it is input into BiLSTM-AT and DPCNN network channels in two channels. In the BiLSTM-AT model channel, perform the global feature acquisition of the secondary context information on the incoming word vector and calculate the weight distribution coefficient through the attention mechanism to give more attention to the key parts; in the DPCNN channel, the incoming text vector Perform deep-level local feature extraction and obtain long text clustering dependencies; finally, the text feature vectors output by the two channels are spliced into the fully connected layer and then the final classification result is calculated by Softmax.

### 3.1. ERNIE

The knowledge-enhanced representation through knowledge integration (ERNIE) [9] is proposed by Baidu, which uses multi-source data and related prior knowledge for pre-training. The formation process of the input vector of the ERNIE model is the same as that of BERT [10]. In the short sentence-level text classification, the sentence is used as the initial input, and the word embedding is encoded as a static word vector in the unit of word. The sentence embedding and the corresponding position embedding are added together as the input of the ERNIE layer. The input vector formation process of the ERNIE layer is shown in Fig. 2 The input vector formation process of the ERNIE model.

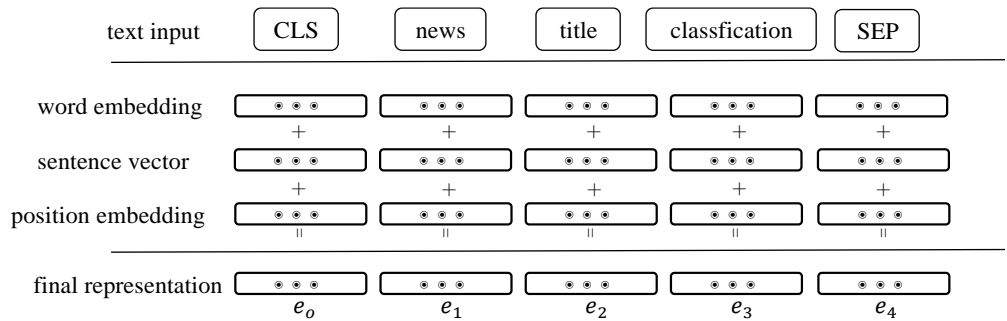

Figure 2. ERNIE model input vector formation process

In Figure.2 above, [CLS] represents a placeholder for the beginning of a sentence, which contains the information of the entire sentence; [SEP] represents a separator used to distinguish different sentences. The static word vector $\{e_0, e_1, e_2, e_3, e_4\}$ is generated by summing the input sentence in the word embedding representation, the whole sentence representation and the position representation vector as the input vector of the "news headline classification" before being passed to the ERNIE layer Characterization. The ERNIE layer extracts the underlying lexical and semantic information from the input vector, and finally generates a dynamic word vector representation that integrates the context.

The granularity of the masking strategy in ERNIE is based on the entity/phrase level. The external knowledge information is not directly input into the model. It is necessary to implicitly learn the knowledge information such as entity relationship and entity attributes, and cover by global information prediction. Content snippets. The concealment strategy is as shown in Figure.3 for a simple illustration of the entity-level concealment in ERNIE.

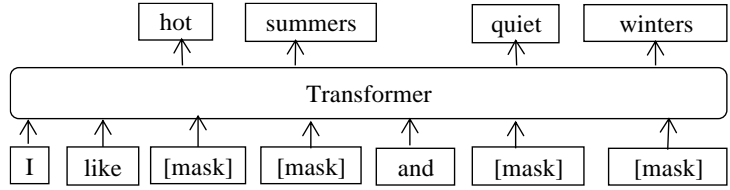

Figure 3. Entity-level masking in ERNIE

The structure design of the ERNIE model is still essentially an encoder based on bidirectional self-attention mechanism. The encoder part comes from the Transformer model and uses a multi-head self-attention mechanism with 12 attention heads. head self-attention) [11], through which to capture the contextual information of the word vector in the text sequence and generate a word vector representation that integrates the context. The relevant formula is calculated as follows.

$$Attention(\boldsymbol{Q}, \boldsymbol{K}, \boldsymbol{V}) = Softmax\left(\frac{\boldsymbol{QK^T}}{\sqrt{d_k}}\right)\boldsymbol{V} \quad (1)$$

$$head_1 = Attention(\boldsymbol{Qw_i^Q}, \boldsymbol{KW_i^K}, \boldsymbol{VW_i^V}) \quad (2)$$

$$MultiHead(\boldsymbol{Q}, \boldsymbol{K}, \boldsymbol{V}) = Concat(head_1, \dots, head_n)\boldsymbol{W^0} \quad (3)$$

Among them, $\boldsymbol{Q}$, $\boldsymbol{K}$, and $\boldsymbol{V}$ represent query matrix, key matrix and value matrix respectively. The function of $\sqrt{d_k}$ is to prevent the inner product from being too large and to overcome the problem of too small gradient that may be caused by high-dimensional matrix operations. $Softmax$ normalizes the calculation result and calculates the weight coefficient of each word. $w_i^Q, W_i^K$, and $W_i^V$ are weight parameter matrices relative to $\boldsymbol{Q}$, $\boldsymbol{K}$, and $\boldsymbol{V}$ respectively.

### 3.2. LSTM and BiLSTM

BiLSTM (bi-drectional long short-term memory, BiLSTM) is a combination of forward long short-term memory (LSTM) and backward LSTM network models. A single LSTM network model can only encode information from front to back. Although it effectively alleviates the long-distance dependence problem of general recurrent neural network (RNN) and the problem of gradient disappearance, it cannot capture the bidirectional semantic features of text vectors. Therefore, using BiLSTM as a channel carrier of the overall model architecture, feature extraction of global contextual semantic information is performed on the input text vector for better feature expression.

The LSTM network model structure is mainly composed of forget gates, inputs, cell state updates, and output gates. It was originally proposed by Hochreiter S and Schmidhuber J [12]. It was the mainstream reference model in the early days of deep learning. The overall network architecture is shown in Figure.4 the simplified structure of the LSTM network model.

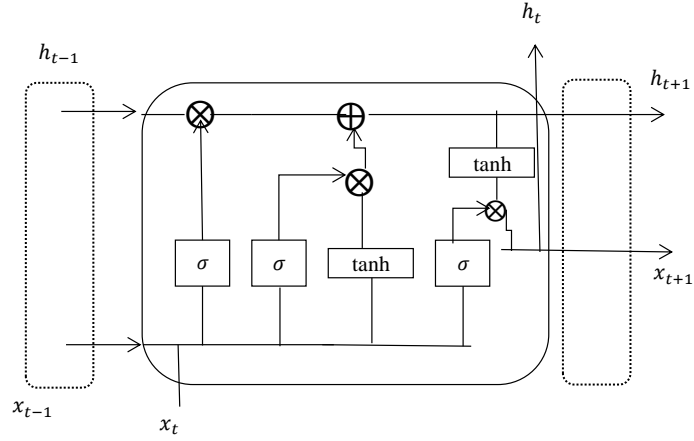

Figure 4. LSTM network model simplified structure.

The activity state between the internal unit structures is shown in formula (4) to formula (9):

$$i_t = \sigma(W_e[h_{t-1}, x_t] + b_i) \quad (4)$$
$$f_t = \sigma(W_f[h_{t-1}, x_t] + b_f) \quad (5)$$
$$\tilde{c}_t = tanh(W_c[h_{t-1}, x_t] + b_c) \quad (6)$$
$$c_t = f_t \odot c_{t-1} + i_t \odot \tilde{c}_t \quad (7)$$
$$o_t = \sigma(W_o[h_{t-1}, x_t] + b_o) \quad (8)$$
$$h_t = o_t \odot tanh(C_t) \quad (9)$$

In the LSTM network model structure, the dashed boxes on the left and right sides represent the structural units of the LSTM at the previous moment and the next moment. In the above formula, $i_t$ represents the input gate, $f_t$ represents the forget gate, and $o_t$ represents the output gate. This is the classic gate control mechanism (Gate mechanism) in the model. $\tilde{c}_t$ represents the cell state, $c_t$ represents the updated cell state, $h_t$ represents the output of the final hidden layer state, $\sigma$ represents the *Sigmoid* activation function, $x_t$ represents the input at the current moment, $b$ represents the bias value, and $\odot$ represents the Hadamard product.

BiLSTM, as a forward and backward combination model of the LSTM network structure, bidirectionally encodes the input text sequence information through the forward and backward network models, fully considering the contextual semantic information, and can extract more accurate global feature vector expressions. The overall structure is shown in Figure 2.3 BiLSTM network model structure.

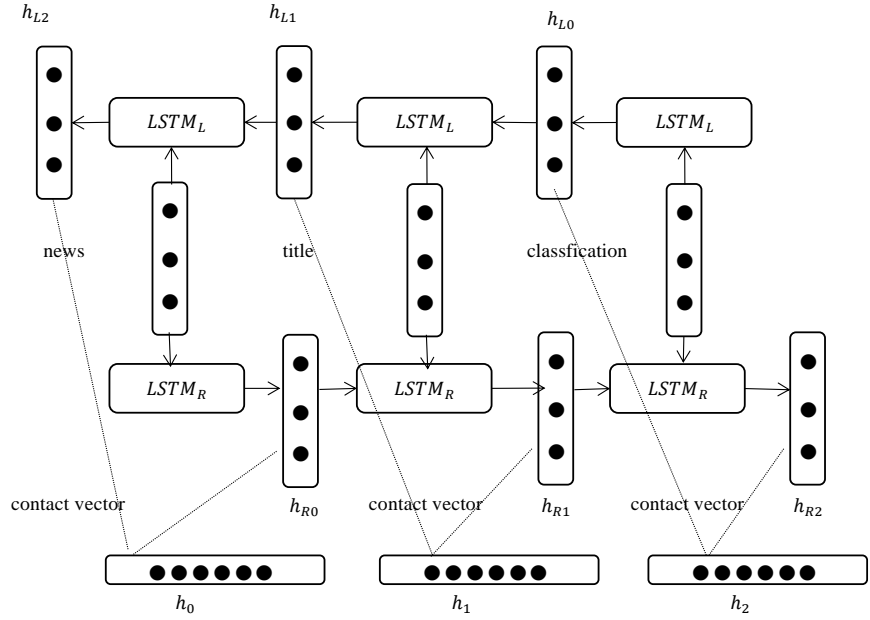

Figure 5. BiLSTM network model structure.

As shown in Figure.5, for the input text sequence "news title classification", the forward input is "news", "title", "category", expressed as a vector $\{h_{R0}, h_{R1}, h_{R2}\}$, and the backward input is "category", "title", "news", expressed as the vector $\{h_{L0}, h_{L1}, h_{L2}\}$. Then the forward and backward hidden layer state vectors are spliced to obtain $\{[h_{R0}, h_{L0}], [h_{R1}, h_{L1}], [h_{R2}, h_{L2}]\}$, that is, the final result is used as the model place .The text vector to be represented by data is $\{h_0, h_1, h_2\}$.

## 3.3. BiLSTM-Attention

After the context global feature extraction of the input text vector is completed by BiLSTM, it is given a higher weight to the key parts that play a decisive role through the attention mechanism layer. In essence, it is a weight distribution mechanism. The larger the weight, the more critical the feature of the text vector and the greater the impact on the final news headline classification result. The formula for calculating the attention distribution of the text vector output by BiLSTM is as follows.

$$u_t = tanh(W_g H_t + b_g) \quad (10)$$
$$a_t = \frac{exp(u_t)}{\sum exp(u_t)} \quad (11)$$
$$R = \sum a_t H_t \quad (12)$$

In the above formula (7)~(9), $W_g$ is the weight parameter, $b_g$ is the bias term, $H_t$ is the hidden layer state vector output, $u_t$ is the hidden layer state vector output at each moment and the weight parameter that the model needs to learn Multiply, add the offset term, and de-linearize the vector expression by the *tanh* function. $a_t$ is the weight coefficient distribution of each vector feature obtained after the *softmax* function is used to calculate the probability distribution of the weight. The characteristic vector of the final attention distribution is expressed as $R$, which is the result of multiplying the output of the state vector at all times and the calculated weight coefficient and weighting the sum.

The bi-directional long-term and short-term memory network BiLSTM-AT combined with the attention mechanism is mainly used in the text to extract the most important semantic information [13], and its overall model structure is shown in Figure 6 BiLSTM-AT model structure.

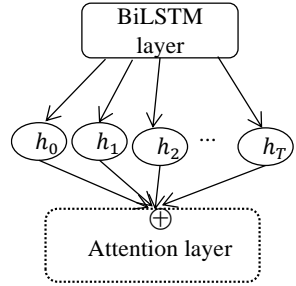

Figure 6. BiLSTM-AT model structure

As shown in Figure.6 above, the feature vector output by the BiLSTM layer is passed to the attention mechanism layer, and the weight coefficient is continuously adjusted through the weighted average to calculate the probability distribution of the attention weight, and the final text feature vector is obtained, which is finally combined with the DPCNN channel The obtained feature vectors are spliced to obtain the final required vector representation.

### 3.4. DPCNN

The deep pyramid convolutional neural network (deep pyramid convolutional neural network, DPCNN) is a word-level deep convolutional neural network. Compared with the traditional convolutional neural network (convolution neural network, CNN) in text classification tasks The application of [14] mainly overcomes the problem of difficulty in extracting long-distance text sequence dependencies [15], and obtains more accurate local features of text vectors through deep convolution while reducing the amount of calculation. Its model structure Figure 7 shows the structure of the DPCNN network model.

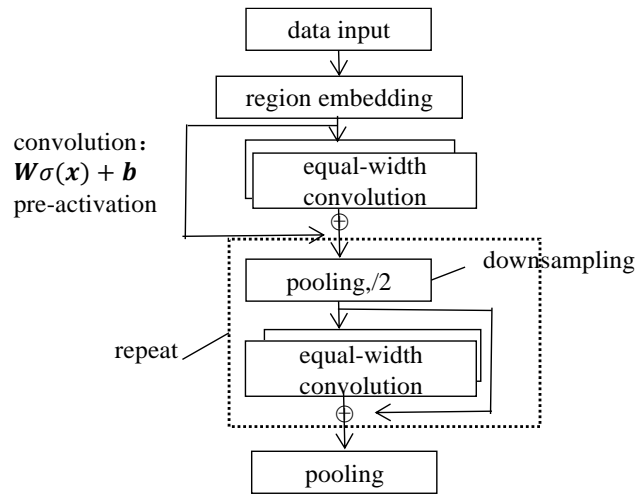

Figure 7. DPCNN network model structure.

As shown in Figure.7 above, Region embedding is essentially a convolutional layer containing convolution filters of different sizes. The input text is convolved here to generate the corresponding word vector encoding. The model contains two convolutional blocks, each convolutional block contains two convolutional layers, and each convolutional layer is equal-width convolution. Assuming that the length of the input sequence is L, the size of the convolution kernel is M, the stride is S, and both ends of the input sequence are filled with N zeros, that is, zero padding. In equal-length convolution, the step size is S=1, the two ends are filled with zero N=(M-1)/2, the output length is still L after convolution, and each layer uses 250 convolution kernels with a size of 3×3 . After each convolution block, a pooling layer with a size of 3 and a step size of 2 is used for 1/2 pooling, that is, downsampling. At the same time, the number of feature maps must be fixed, so the text sequence length is compressed to 1/2 of the original The small fragments of text that can be perceived can be doubled to capture the dependence of long-distance text. And because the size of the text sequence data is halved, the corresponding calculation time will also be halved, thus forming a pyramid structure model. At the same time, use shortcut connect[16] to train the deep learning network, as shown in formula (10), where $z$ is the input vector of the neural network, $f$ stands for skipped layers, and $f(z)$ stands for The two-layer equal-length convolution of pre-activation greatly alleviates the problem of gradient dispersion, thereby completing the learning of identity mapping. The pre-activation method is as shown in formula (11).

$$f(z)+z \quad (13)$$
$$conv: \boldsymbol{W}\sigma(\boldsymbol{x}) + \boldsymbol{b} \quad (14)$$

That is, when performing a convolution operation in a neural network, directly perform the linear operation of the activation function on the input text vector and then multiply it with the weight parameter $\boldsymbol{W}$ and add it to the offset term $\boldsymbol{b}$, instead of performing the activation function on the entire already-calculated variable Linear conversion. The activation function $\sigma$ in DPCNN represents the *Relu* activation function.

## 4. EXPERIMENTAL PROCESS AND RESEARCH

### 4.1. Experimental Data

The news headline classification data set used in the experiment used Tsinghua University's Chinese text classification toolkit THHUCTC[17], and 200,000 news headline data were extracted from the Chinese text classification data set THUCNews, including: finance, real estate , Stocks, education, technology, society, current affairs, sports, games, entertainment ten categories. Each category has a total of 20,000 data, and the length of the text sequence is within 30 words.

### 4.2. Experimental environment and design

The 200,000 Chinese news headline data sets used in the experiment are set to 180,000 in the training set, 10,000 in the validation set, and 10,000 in the test set. The three data sets contain ten categories, and the number of each category is evenly distributed. , That is, there are 18,000 items per category in the training set, 1,000 items per category in the validation set, and 1,000 items per category in the test set.

In the experiment of the news headline classification task, the model parameters are set as shown in Table 1. The stop_go variable indicates that if the neural network model is training and learning, if the experimental effect of more than 1000 batch_size is not improved, the training will be terminated and the result will be output.

Table 1. DC-EBAD Model Parameters

| Model Channel | Parameters | Value |
| --- | --- | --- |
| ERNIE Layer | hidden_size | 768 |
| BiLSTM-AT-Channel | r_hidden_size | 256 |
|  | num_layers | 2 |
| DPCNN-Channel | num_filters | 250 |
|  | Kernel_size | 3 |
| Hyperparameter | batch_size | 128 |
|  | text_size | 32 |
|  | dropout | 0.5 |
|  | learning_rate | 5e-5 |
|  | stop_go | 1000 |
|  | epoch | 3 |

In order to verify the advantages of the proposed model DC-EBAD, single-channel ERN-IE-BiLSTM-AT and ERNIE-DPCNN models us-ing pre-trained language model layers, two-channel BiLSTM-AT-DPCNN models without p-re-training models and other neural networks -are set up Models LSTM, BiLSTM, BiLSTM+Attention, DPCNN, CNN, etc. are used as comparative experiments, and experiments are carried out in the same data set and experiment-al environment.

### 4.3. Experimental Evaluation Index

The evaluation indicators of news headline classification used in the experiment are accuracy and F1-score. The accuracy rate is to evaluate the classification accuracy of the ten news headline categories as a whole, and the F1-score is for each of the ten categories.

The formula for accuracy, precision, recall and F1-score is defined as follows:

$$accuracy = \frac{TP + TN}{TP + TN + FP + FN} \quad (15)$$

$$precision = \frac{TP}{TP + FP} \quad (16)$$

$$recall = \frac{TP}{TP + FN} \quad (17)$$

$$F1 = \frac{2 * precision * recall}{precision + recall} \quad (18)$$

In the above formula, TP means that the label is true and the prediction result is true; TN means that the label is true and the prediction result is false; FP means the label is false and the prediction result is true, and FN means the label is false and the prediction result is false.

### 4.4. Analysis of Experimental Results

The experimental evaluation indicators are evaluated from the overall accuracy, precision and F1-score of the data set, as well as the F1-score of each sub-category. For the neural network models ①~⑥ that do not use the ERNIE pre-trained language model, pre-training word vectors are used to construct the vocabulary of the input text data and generate the corresponding word embedding matrix to complete the vectorized representation of the text. Here the pre-training word vector is the word/word pre-training word vector of Sogou News [18]. In the experiment, the input of text data in units of words is completed, and the first line of the pre-training word vector is the word vector File information description: "365076 300", the first number is 365076, that is, the number of words/word vectors contained in the text, and the second indicates that the dimension of the word vector embedding is 300 dimensions. Each subsequent row includes a character/word and its 300-dimensional vector representation.

Table 2. Accuracy of experimental results

| Number | Model | Accuracy | Precision | F1-score |
|---|---|---|---|---|
| ① | LSTM | 90.69% | 90.69% | 90.67% |
| ② | CNN | 90.74% | 90.77% | 90.74% |
| ③ | DPCNN | 91.43% | 91.51% | 91.45% |
| ④ | BiLSTM | 90.98% | 91.07% | 91.01% |
| ⑤ | BiLSTM-AT | 91.44% | 91.49% | 91.44% |
| ⑥ | **BiLSTM-AT-DPCNN** | **92.02%** | **92.08%** | **92.03%** |
| ⑦ | ERNIE-BiLSTM-AT | 94.56% | 94.58% | 94.56% |
| ⑧ | ERNIE-DPCNN | 94.58% | 94.63% | 94.59% |
| ⑨ | **DC-EBAD** | **94.63%** | **94.66%** | **94.63%** |

It can be seen from Table 2 that the conventional CNN and LSTM neural network models performed relatively well on the Chinese news headline data set used in the experiment, with little difference, exceeding 90%. DPCNN, BiLSTM and BiLSTM-AT, as their theoretical optimization models relative to CNN, LSTM and BiLSTM, have also improved various indicators in actual performance. BiLSTM-AT-DPCNN is the dual-channel parallel network model after removing the ERNIE pre-trained language model layer in the DC-EBAD model. Compared with other single-channel neural network comparison models, it is the best and surpasses all three indicators. That's 92%.

After using ERNIE as the pre-training language model layer, the indicators of the three models ⑦~⑨ have reached more than 94%, and the DC-EBAD model has the best performance, surpassing 94.60%, which is more accurate than without ERNIE The dual-channel model BiLSTM-AT-DPCNN of the pre-trained language model layer increased by 2.61%. Compared with the other two single-channel models ERNIE-BiLSTM-AT and ERNIE-DPCNN that use the ERNIE pre-training model, it has increased by 0.07% and 0.05%, respectively.

However, after using the pre-trained language model to represent the word vector of the input text, there will be certain requirements for the hardware facilities of the experimental environment. In this experiment, a single 12G-GPU device is used. In the case of 200,000 pieces of data, The training and learning time of the model has been greatly increased, with ERNIE-DPCNN and DC-EBAD both exceeding 5 hours, while ERNIE-BiLSTM-AT running time exceeds 6 hours.

① ~⑨ in Table III are the corresponding serial numbers of each network model in Table 2.

Table 3. F1-score of each category of experimental results

| Category | F1-① | F1-② | F1-③ | F1-④ | F1-⑤ | **F1-⑥** | F1-⑦ | F1-⑧ | **F1-⑨** |
|---|---|---|---|---|---|---|---|---|---|
| finance | 89.92% | 90.59% | 90.57% | 90.76% | **91.10%** | 90.89% | **93.63%** | 93.09% | 93.36% |
| estate | 91.43% | 92.41% | 92.80% | 91.93% | 92.39% | **92.89%** | 95.69% | **95.81%** | 95.68% |
| **stock** | 84.50% | 85.57% | 85.48% | 84.68% | 85.57% | **86.55%** | **90.61%** | 90.16% | 90.08% |
| education | 93.81% | 95.14% | 94.43% | 94.75% | 94.95% | **95.14%** | 96.78% | 96.82% | **96.92%** |
| **technology** | 85.37% | 86.56% | 86.92% | 84.88% | 87.39% | **88.15%** | 91.39% | 91.12% | **91.83%** |
| society | 90.73% | 88.92% | 91.57% | 90.76% | 90.52% | **91.50%** | 94.14% | 94.39% | **94.73%** |
| politics | 87.87% | 89.30% | 89.67% | 88.07% | 88.66% | **89.74%** | 92.79% | **93.22%** | 92.26% |
| sport | 97.08% | 95.43% | 96.88% | **97.73%** | 97.01% | 97.65% | 98.40% | **99.00%** | 97.97% |
| game | 93.48% | 91.41% | 93.50% | 93.24% | 94.13% | **94.29%** | 96.19% | 96.00% | **96.68%** |
| entertainment | 92.51% | 92.07% | 92.68% | 93.31% | 92.70% | **93.54%** | 95.96% | 96.25% | **96.79%** |

It can be seen from Table 3 that when the usual deep neural network ①~⑥ is used to classify news headlines, the BiLSTM-AT-DPCNN model is slightly inferior to other models in terms of financial and sports classification effects. The performance in the remaining 8 categories is the best. However, it can be seen that the highest F1-score of all models in the stock and technology data sets are 86.55% and 88.15%, respectively. Compared with the highest F1-score of other categories, it can be seen that the classification of the two types of news headline text data is important for the model. It is still somewhat confusing, and the model is still lacking in data feature extraction and learning.

After using the pre-trained language model ERNIE combined with the dual-channel network model, the highest levels of stock and technology category classification reached 90.61% and 91.83%, respectively. Compared with the best-performing dual-channel model BiLSTM- when the pre-trained model was not used, The index on AT-DPCNN increased by 4.06% and 3.68% respectively. The best performance in finance and stocks is the ERNIE-BiLSTM-AT model, and the best performance in real estate, current affairs and sports is the ERNIE-DPCNN model, and its F1-score in sports is as high as 99.00%. Compared with the two single-channel comparison models ERNIE-BiLSTM-AT and ERNIE-DPCNN, the DC-EBAD model has the best performance in the categories of news headlines in education, technology, society, games, and entertainment. The performance of the proposed DC-EBAD Chinese news headline classification model is considerable.

## 5. CONCLUSION

The DC-EBAD dual-channel model based on the ERNIE pre-training model proposed in the article better realizes the classification of large-scale multi-category news headline data. The pre-training language model is used to extract the lexical and semantic information of the underlying text, and generate input text fusion The dynamic word vector representation of the context information is passed into the dual-channel network to extract the key parts of the global feature and the deep-level local feature extraction respectively, and the splicing vector obtains the final feature vector expression. The evaluation indicators of the experimental results are better than the set comparison model, and most of the F1-score of the sub-class results are also improved. In the future work, considering the optimization of the data input part, such as combining the features of part of speech and other features to further enhance the representation information, so that the neural network can better perform feature learning, and can achieve higher levels in multiple types of news headline tasks.